\pdftrailerid{}
\documentclass{teleai}

\usepackage{amsmath,amssymb}
\usepackage{booktabs,graphicx,natbib}
\usepackage{needspace}
\setcitestyle{authoryear,round,citesep={;},aysep={,},yysep={;}}
\hypersetup{
    pdftitle={CoCode-World: Shared 3D Proxy Representation for Multi-Player World Models},
    pdfauthor={Qirui Li; Xin Zhang; Yabo Chen; Jinkun Hao; Paul L Rosin; Yu-Kun Lai; Haibin Huang; Yin Chen; Ran Yi; Chi Zhang}
}

\title{CoCode-World: Shared 3D Proxy Representation for Multi-Player World Models}
\author[1,2]{Qirui Li}
\author[2]{Xin Zhang}
\author[2]{Yabo Chen}
\author[1]{Jinkun Hao}
\author[3]{Paul~L~Rosin}
\author[3]{Yu-Kun~Lai}
\author[2]{Haibin Huang}
\author[2]{Yin Chen}
\author[1,*]{Ran Yi}
\author[2,*]{Chi Zhang}

\affiliation[1]{Shanghai Jiao Tong University}
\affiliation[2]{Institute of Artificial Intelligence, China Telecom (TeleAI)}
\affiliation[3]{Cardiff University}
\contribution[*]{Corresponding authors.}

\abstract{Multiplayer generative world models must produce observations that agree on shared entities despite different camera motions and occlusions.
Explicit world representations provide spatial control, but precise modeling of individual game maps limits scene diversity, while shared geometry alone cannot ensure consistent appearance across views.
We present CoCode-World, a shared 3D proxy framework that separates structural control from visual synthesis for two-player video generation.
Programmable proxies specify scene layout, motion, and occlusion, while RGB references guide character and scene appearance.
To coordinate visual details across players, we concatenate paired proxy videos, appearance references, and target views in a consistent spatial layout and generate both views jointly with a pretrained video model.
This formulation uses the model's native attention for cross-view communication within a single generation branch, with LoRA adaptation learning proxy adherence and appearance correspondence from paired observations.
We construct a two-player GTA~V dataset that pairs synchronized RGB observations with proxy videos derived from shared map geometry and game-engine state.
Experiments demonstrate appearance transfer to characters and scene styles unseen during training, with improvements in video quality and cross-view player identity over the evaluated baseline.
Ablations further examine how proxy construction and cross-view communication affect generation.
Project page: \url{https://iapple233.github.io/CCW/}.
}

\begin{document}
\maketitle

\begin{figure}[!ht]
    \centering
    \includegraphics[width=\linewidth]{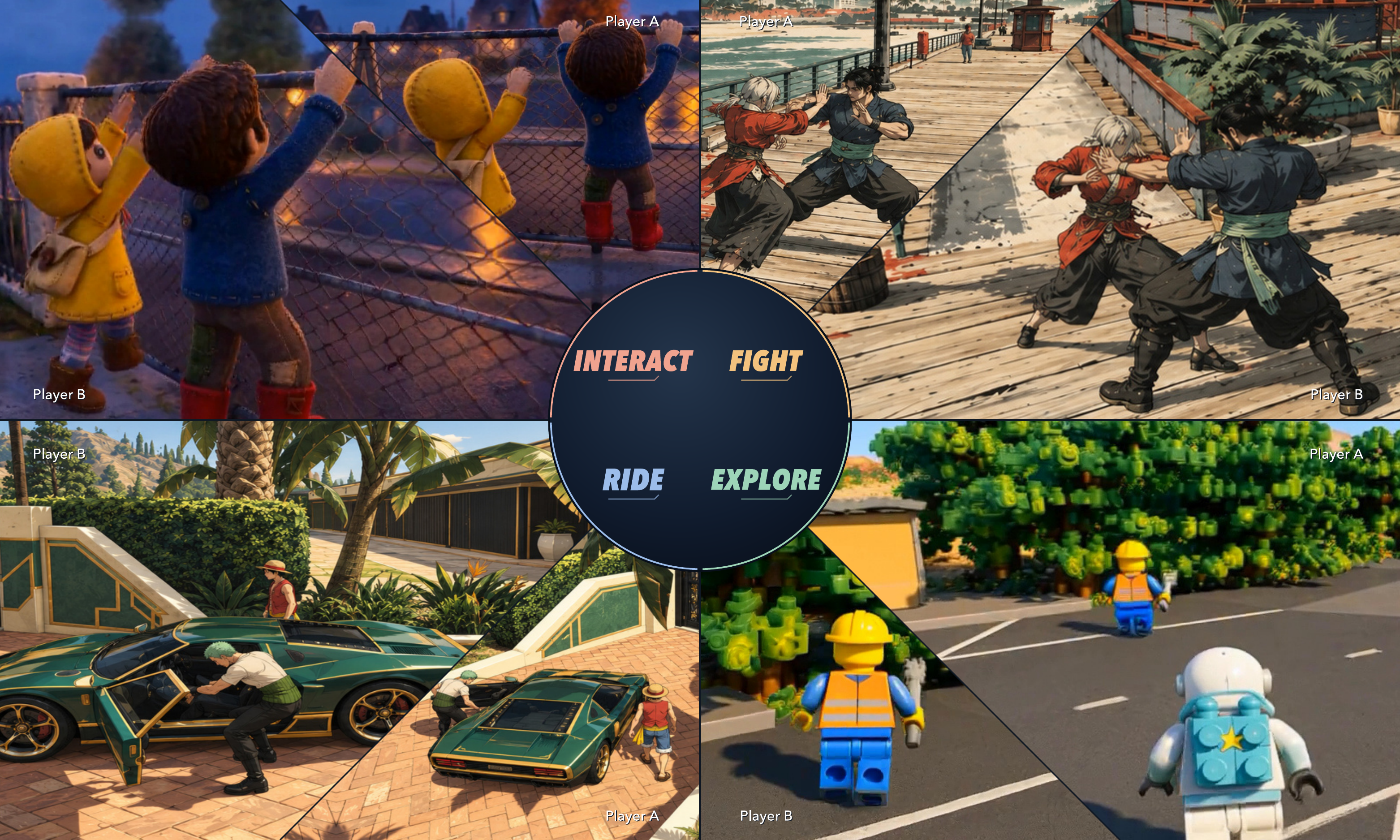}
    \caption{\textbf{Two-player views of a shared game world.} Paired views illustrate environment interaction, combat, exploration, and vehicle riding. Each pair shows the scene from the two players' perspectives.}
    \label{fig:teaser}
\end{figure}

\section{Introduction}
\label{sec:introduction}

Generative world models synthesize visual observations in response to interaction, extending video generation toward playable environments~\citep{genie,gamengen,worldplay}.
In a multiplayer setting, each player has a different camera but participates in the same scene.
A character seen from behind by one player may be visible from the front to another, and an object leaving one field of view may remain visible elsewhere.
The generated observations must agree about these shared entities while following different camera motions and occlusions.
This requirement couples spatial control with appearance consistency across views.

Existing multiplayer world models often coordinate player observations in learned latent spaces through joint denoising, cross-agent attention, or shared representations~\citep{solaris,shareverse,gamma,multiworld}.
Recently, explicit-state methods show the potential of separating world evolution from view synthesis~\citep{mass}.
However, some methods rely on detailed, scene-specific map geometry during training~\citep{worldcast}, complicating transfer to different scenes and limiting the diversity of playable environments.
We therefore explore coarse, programmable 3D proxies through which a coding agent constructs and updates a world, while a video model supplies its visual details.

Recent work on video models as renderers supports this separation of structural control and visual synthesis.
Code World Model~\citep{cwm} translates executable world state into proxy videos, while SolarRenderer~\citep{fang2026solarrendererfewstepgenerativereference} synthesizes detailed video from structural and appearance references.
These works primarily study single-view synthesis.
Extending this formulation to multiplayer environments introduces a further challenge: a shared proxy constrains scene layout, motion, and occlusion, but does not determine how the same entity should appear across player views.
Independently generated views can therefore follow their respective proxies yet assign different clothing or textures to a shared character.
Coordinating these visual choices calls for a communication mechanism that exchanges appearance information across views during generation.

To address this challenge, we propose CoCode-World, a shared 3D proxy framework with cross-view communication for joint two-player video generation.
Prior approaches such as MetaWorld~\citep{metaworld} use separate generation branches for different players and add cross-attention modules for communication.
We build on the in-context learning capability of pretrained video models to generate both views within a single branch.
Specifically, we adopt the MiniMax-H3 33B Ref2VA backbone~\citep{minimaxh3} and concatenate the two views along image width, using the same ordering for the proxy videos, appearance references, and target videos.
The native attention layers allow each view to access appearance information from both players throughout denoising.
We keep the pretrained backbone frozen and train LoRA adapters on paired observations to learn the correspondence between proxy motion and rendered video, as well as appearance consistency across views.

We instantiate the framework using two-player GTA~V recordings, exported map geometry, and per-client camera and entity observations.
The resulting dataset contains 7,846 paired training clips, each with synchronized RGB observations and proxy videos rendered from the two player cameras.
Shared map geometry and persistent entity identifiers connect the views in a common coordinate system, while the game engine supplies the trajectories and state observations.
The proxies constrain layout and motion, and RGB references specify the desired appearance, allowing the video model to render the same gameplay with different characters and scene styles.

We evaluate generalization on a benchmark of 150 cases, each featuring two distinct characters in a shared scene.
Although trained on GTA~V imagery, CoCode-World learns to follow proxy motion while generating character and scene appearances unseen during training.
Our ablations examine how proxy construction affects generation and show that joint generation better preserves character details across player views than a dual-branch variant with cross-attention.

\begin{itemize}
    \interlinepenalty=10000
    \item We propose CoCode-World, a shared 3D proxy framework for controllable two-player video generation with flexible character and scene appearance.
    \item We develop a joint generation method that uses the native attention of a pretrained video model for cross-view communication and better preserves character appearance than a dual-branch alternative.
    \item We construct a two-player GTA~V dataset with 7,846 paired clips, combining synchronized RGB observations with proxy videos derived from game-engine geometry and state. The dataset also includes the shared 3D world representation, world states, and player-action metadata, enabling the community to build proxies tailored to their needs.
    \item We achieve state-of-the-art results among the evaluated methods in cross-view player identity and video quality, outperforming Seedance~2.5 and SolarRenderer.
\end{itemize}

\section{Related Work}
\label{sec:related_work}

\paragraph{Interactive Generation and Memory.}
Video generation now covers joint audio-video synthesis and long-form, multi-shot narratives~\citep{evolution,harmony,cinedance}.
Genie learns controllable environments from videos through latent actions~\citep{genie}, and GameNGen predicts gameplay observations from previous frames and actions~\citep{gamengen}.
Maintaining a coherent environment over longer interactions requires information beyond the current visual context.
WorldMem retrieves historical observations using their associated states~\citep{worldmem}, VMem indexes views by the surface elements they observe~\citep{vmem}, and TeleWorld couples generation with 4D reconstruction and persistent memory~\citep{chen2025teleworlddynamicmultimodalsynthesis}.
WorldPlay studies geometric consistency in real-time generation~\citep{worldplay}, and SolarWM examines data and training for long-horizon world models~\citep{huang2026solarwmopendatascalable}.
Cycle-World reduces error accumulation through reverse-prediction cycle consistency~\citep{cycleworld}.
Our setting adds a spatial relation between simultaneous player observations: the same entity can be visible through different cameras within one generated interval.

\paragraph{Multiplayer and Multiview Generation.}
Solaris jointly models synchronized Minecraft observations and actions~\citep{solaris}.
ShareVerse exchanges features through cross-agent attention~\citep{shareverse}, and Gamma-World uses permutation-symmetric encoding and sparse hub attention to coordinate multiple agents~\citep{gamma}.
Geometric representations provide another source of cross-view information.
MultiWorld extracts a 3D-aware global latent state from multiple observations~\citep{multiworld}; Prisma-World combines relative camera geometry, joint attention, and minimap guidance~\citep{prisma}; MetaWorld aligns camera and subject trajectories across generation branches~\citep{metaworld}.
Our renderer also couples the generation process, using camera-specific projections of a shared 3D scene as its structural conditions.

\paragraph{Explicit State and Visual Synthesis.}
MultiGen separates memory, observation, and dynamics, using an external level map and player poses to support multiplayer rollouts~\citep{multigen}.
MASS combines a logic engine that updates typed state with a rendering engine that produces camera-specific observations~\citep{mass}, while PhysOmni and PhysAgent decouples physics simulation from visual synthesis~\citep{zhang2026physomniphysicsgroundedmultiobjectscene, li2026physagent}. EgoSim generates action-conditioned observations and updates a persistent 3D scene state from them~\citep{hao2026egosim}. 
These methods separate world-state maintenance from appearance synthesis.
We adopt this separation at the rendering interface: an external environment supplies scene and entity observations, and a joint video model translates their projections into player views.
Our method addresses conditional observation synthesis; learning action-driven state transitions is outside the renderer.

\paragraph{Structural Conditions and Proxy Rendering.}
ControlNet demonstrates spatial control of pretrained diffusion models using conditions such as depth and pose~\citep{controlnet}.
PoseAnything extends pose guidance to human and non-human subjects~\citep{poseanything}, while MotionMaster transfers camera motion without training~\citep{motionmaster}.
Related controls include camera and object trajectories in SymphoMotion~\citep{Zhang_2026_CVPR}, geometry latents in VideoWeave~\citep{xiang2026videoweaveunlockinggeometricconsistency}, and depth trajectories with reference appearance in UniWorld~\citep{zhang2026uniworldunifiedworldmodel}.
Code World Model maintains executable state and compiles it into coarse proxy video~\citep{cwm}, while ProxyUp uses proxy dynamics to guide generated content~\citep{proxyup}.
Seedance~2.5 also supports textureless 3D render references~\citep{seedance25}.
SolarRenderer adapts native reference conditioning for video synthesis from structural and appearance inputs~\citep{fang2026solarrendererfewstepgenerativereference}.
We extend this rendering formulation to paired observations of a common scene.
The structural inputs connect the cameras through map coordinates and entity identities, and the concatenated generation context makes appearance evidence available across views.

\section{Method}
\label{sec:method}

\begin{figure}[!htbp]
    \centering
    \includegraphics[width=\linewidth]{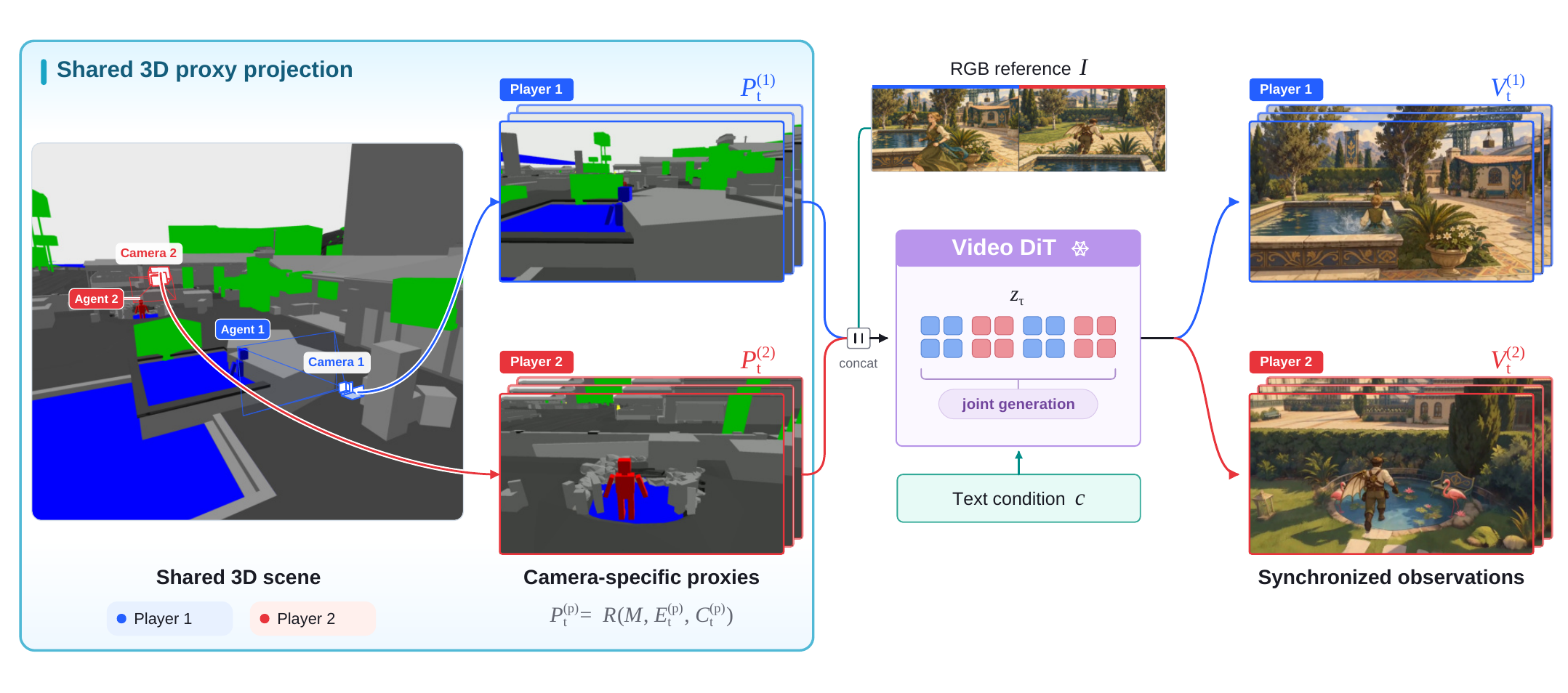}
    \caption{Joint two-player rendering. Map geometry and entity observations produce a proxy for each camera. The two proxy views share a fixed left--right layout with the RGB appearance reference and target video. The H3 transformer jointly denoises both target views through its native reference interface, with full attention across the context and LoRA adaptation of the frozen backbone.}
    \label{fig:method_overview}
\end{figure}

Given two camera trajectories, their proxy videos, and a paired appearance reference, our goal is to generate both players' observations of the same scene.
Figure~\ref{fig:method_overview} summarizes the renderer.
We first construct paired structural observations from gameplay, then organize the two views in a shared generation context and adapt a pretrained video transformer to this task.

\subsection{Dataset Construction and 3D Proxy--Video Alignment}
\label{sec:multiplayer_data}

\paragraph{Scene Representation.}
We collect two-player GTA~V gameplay through a private FiveM server.
For each client, we record RGB video, camera parameters, and the identities and poses of players and nearby entities.
We export static map geometry in the same coordinate system.
The map and persistent entity identifiers connect observations of the same scene across cameras.
Dynamic observations retain the states available to each client under local streaming and network updates.

Let $\mathcal{M}$ denote the shared map, $\mathcal{E}_t^{(p)}$ the entity observations of player $p$, and $C_t^{(p)}$ its camera pose and projection at time $t$.
The proxy renderer produces
\begin{equation}
    P_t^{(p)}=\mathcal{R}(\mathcal{M},\mathcal{E}_t^{(p)},C_t^{(p)}),
    \qquad p\in\{1,2\}.
    \label{eq:proxy_render}
\end{equation}
The map preserves surfaces and obstacles, and simplified character and vehicle geometry represents dynamic entities.
Projecting these elements into each camera makes their relative positions, scale, motion, and occlusions available as image-space conditions.
The proxy is textureless: its category colors identify coarse entities, while clothing, materials, and scene appearance come from the RGB reference and video prior.

\begin{figure}[!htbp]
    \centering
    \includegraphics[width=\linewidth]{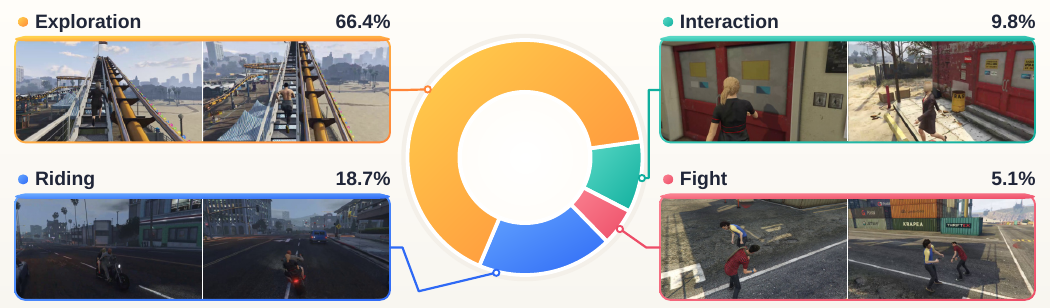}
    \caption{\textbf{Gameplay categories and examples.}
    The donut shows clip proportions across four categories, with colored connectors linking each category to paired gameplay views.
    Each of the 7,846 approximately five-second clips receives one primary category from caption-described temporal coverage across both views; ties follow fight, riding, interaction, then exploration.
    Proportions are automatic caption-derived estimates.}
    \label{fig:dataset_overview}
\end{figure}

\paragraph{Paired Supervision.}
We align the two recordings using estimated server timestamps and retain intervals with observations from both players.
Each RGB frame is paired with a proxy rendered from its associated camera and entity observations.
A training example contains the two RGB video clips, two proxy clips, a joint appearance reference image, and a caption.
We take the first RGB frame of each view as its training appearance reference.
Captions describe the scene, shared character identities, and the actions and visibility within each view.
The paired targets supervise both proxy adherence within a camera and the appearance of shared entities across cameras.
Figure~\ref{fig:dataset_overview} summarizes the clip distribution across four gameplay categories and shows paired visual examples.

\subsection{Joint Generation in a Shared Context}
\label{sec:joint_generation}

\paragraph{Spatial Organization.}
The proxy specifies geometry but leaves visual detail ambiguous.
To make appearance evidence available across cameras, we concatenate the views along image width:
\begin{equation}
    V_t=[V_t^{(1)}\mid V_t^{(2)}],\qquad
    P_t=[P_t^{(1)}\mid P_t^{(2)}],
    \label{eq:spatial_packing}
\end{equation}
where $V_t^{(p)}$ is an RGB frame.
Player~1 always occupies the left half and player~2 the right half, including in the joint reference $I$.
With $\mathbf{V}=(V_t)_{t=0}^{T-1}$, $\mathbf{P}=(P_t)_{t=0}^{T-1}$, and caption $c$, we model
\begin{equation}
    p_{\theta,\phi}(\mathbf{V}\mid\mathbf{P},I,c),
    \label{eq:joint_distribution}
\end{equation}
where $\theta$ denotes frozen pretrained parameters and $\phi$ the learned adaptation.
The two halves preserve their camera-specific content within one prediction task.

\paragraph{Reference Conditioning.}
We use the native image-and-video reference interface of MiniMax~H3.
The joint RGB image provides appearance evidence, and the joint proxy video specifies the evolving structure.
The backbone encodes these references and the target video into a shared attention context.
Full bidirectional attention lets a target region use reference and target features from either player.
For example, clothing visible in one reference view can inform synthesis of the same character in the other view.
The caption supplies scene and action information that complements the visual conditions.

The reference segments remain conditioned while the target is denoised.
We retain the backbone's native reference encoding and attention layers, with LoRA applied to the attention projections and feed-forward layers~\citep{hu2021loralowrankadaptationlarge}.
Unlike treating each view as a separate training example, the paired canvas exposes both observations to every joint prediction.
Cross-view appearance correspondence is learned from the paired RGB supervision; the geometric representation supplies the camera-specific structural constraints.

\subsection{Learning and Inference}
\label{sec:learning}

We use the backbone's flow-matching objective to adapt the renderer~\citep{huang2026solarwmopendatascalable}.
Let $z$ be the clean latent of the joint RGB clip and $\epsilon\sim\mathcal{N}(0,\mathbf{I})$ Gaussian noise.
For flow time $\tau\in[0,1]$, the noised target is $z_\tau=\tau z+(1-\tau)\epsilon$.
We optimize the task adapters with
\begin{equation}
    \mathcal{L}(\phi)=\mathbb{E}\!\left[
    \frac{1}{N}\left\|
    v_{\theta,\phi}(z_\tau,\tau;\mathbf{P},I,c)-(z-\epsilon)
    \right\|_2^2\right],
    \label{eq:joint_flow_loss}
\end{equation}
where $v_{\theta,\phi}$ predicts latent velocity and $N$ counts the target latent elements.
The loss covers both player regions and excludes the reference segments.
The expectation is over training examples, noise, and sampled flow times.

At inference, the proxy views and appearance reference use the same layout as in training.
The reference can specify a visual style different from the source gameplay while the proxy retains its trajectories.
We jointly denoise and decode the target canvas, remove spatial padding, and split it into two player videos.
The generation process consumes the supplied proxy interval, so each output is conditioned on the recorded motion throughout that interval.

\section{Experiments}
\label{sec:experiments}

We evaluate how well our renderer translates shared gameplay into visually coherent player views across different appearances and environments.
We first introduce the benchmark and evaluation protocol, then compare with video generation and rendering baselines, followed by qualitative comparisons and ablations of the proxy representation and generation architecture.

\subsection{Experimental Setup}
\label{sec:eval_protocol}

\paragraph{Benchmark.}
We construct a benchmark of 150 two-player cases covering four event categories: exploration, vehicle riding, player fighting, and environment interactions.
The proxy trajectories are sampled from GTA~V gameplay and provide synchronized structural conditions for the two player views.
We use GPT Image~2.5~\citep{openai2026gptimage25} to stylize the corresponding reference images, changing player appearances, background environments, and visual styles.
Each case pairs these appearance references with the original proxy trajectories, testing whether the renderer can preserve the shared gameplay under appearance and scene changes.

\paragraph{Evaluation Protocol.}
Following SolarRenderer~\citep{fang2026solarrendererfewstepgenerativereference}, we assess visual quality, temporal stability, structural adherence, and reference preservation.
DOVER~\citep{wu2023exploringvideoqualityassessment} and the LAION aesthetic predictor measure video quality and frame aesthetics, respectively.
DINO Adj. measures temporal appearance stability through the cosine similarity of adjacent-frame DINOv2 embeddings~\citep{oquab2024dinov2learningrobustvisual}.
Edge F1 measures agreement between the generated and proxy contours, while First-frame measures DINOv2 similarity between the final second of the generated video and the supplied appearance reference.
We evaluate both player views over the same temporal interval, average the view scores within each case, and then average across cases.

We additionally evaluate whether the same player retains a consistent identity and action across views.
The proxy provides player identities, visibility, and spatial correspondences at each second.
Whenever a player is visible in multiple views, we use these correspondences to extract synchronized crops of that player from the generated videos.
Qwen3.8-27B compares each crop pair and separately judges whether the depicted player identity and action agree, accounting for changes in viewpoint.
The paired judgments test player consistency across simultaneous observations.

\Needspace{8\baselineskip}
\subsection{Quantitative and Qualitative Comparison}
\label{sec:quantitative}
\label{sec:seedance_comparison}

\paragraph{Baselines.}
Some multiplayer world models rely on detailed, scene-specific map geometry during training~\citep{worldcast}, making direct comparison across the diverse scenes in our benchmark difficult.
We therefore select Gamma-World~\citep{gamma} as a multiplayer baseline: it generates player observations conditioned on initial frames and player action sequences, without requiring an explicit scene map as input.
We also compare with two general-purpose video models, MiniMax-H3~\citep{minimaxh3} and Seedance~2.5~\citep{seedance25}, each conditioned on the corresponding appearance references and proxy trajectories.
SolarRenderer~\citep{fang2026solarrendererfewstepgenerativereference} provides a further baseline that uses a video model as a renderer.
Since SolarRenderer supports a single player view, we generate the two views separately from their respective proxies and appearance references.
All methods are evaluated on the same benchmark using the same evaluation protocol.
To examine an existing multiplayer generation strategy on the same backbone, we adapt MetaWorld's dual-branch alignment mechanism from its Wan backbone to MiniMax-H3 and fine-tune it on our paired data; Section~\ref{sec:generation_ablation} presents this comparison.

\begin{table}[!htbp]
    \centering
    \caption{Quantitative comparison. The first five metrics follow SolarRenderer: visual quality (DOVER and LAION), temporal appearance stability (DINO Adj.), proxy contour agreement (Edge F1), and reference preservation (First-frame). Cross-view reports the player-identity score on a 0--5 scale. Higher is better for every metric.}
    \label{tab:fullview-results}
    \small
    \setlength{\tabcolsep}{4pt}
    \renewcommand{\arraystretch}{1.15}
    \begin{tabular*}{\linewidth}{@{\extracolsep{\fill}}lrrrrrr@{}}
        \toprule
        Method & DOVER $\uparrow$ & LAION $\uparrow$ & DINO Adj. $\uparrow$ & Edge F1 $\uparrow$ & First-frame $\uparrow$ & Cross-view $\uparrow$ \\
        \midrule
        Gamma-World & 0.170 & 4.364 & 0.923 & 0.038 & 0.160 & 0.697 \\
        SolarRenderer & 0.454 & 5.477 & 0.968 & 0.073 & 0.655 & 3.696 \\
        MiniMax-H3 & 0.441 & 5.397 & 0.958 & 0.060 & 0.659 & 2.868 \\
        Seedance~2.5 & 0.469 & 5.568 & \textbf{0.973} & 0.120 & \textbf{0.700} & 3.552 \\
        Ours & \textbf{0.524} & \textbf{5.655} & 0.970 & \textbf{0.124} & 0.684 & \textbf{3.847} \\
        \bottomrule
    \end{tabular*}
\end{table}

\paragraph{Results.}
Table~\ref{tab:fullview-results} compares visual quality, structural adherence, appearance preservation, and cross-view player identity.
Gamma-World performs substantially worse than the video generation and rendering baselines, with particularly weak reference preservation and cross-view identity consistency.
This result highlights the difficulty of transferring existing multiplayer world models to the diverse scenes and appearances in our benchmark.
Relative to Seedance~2.5, our renderer improves visual quality and agreement with the proxy structure, together with a higher cross-view identity score.
The improvement in identity consistency is particularly relevant to multiplayer rendering, where a player's appearance must agree across simultaneous observations.
Our method also achieves competitive temporal appearance stability and reference preservation compared with Seedance~2.5.

\setcounter{topnumber}{1}
\begin{figure}[t]
    \centering
    \includegraphics[width=0.90\linewidth]{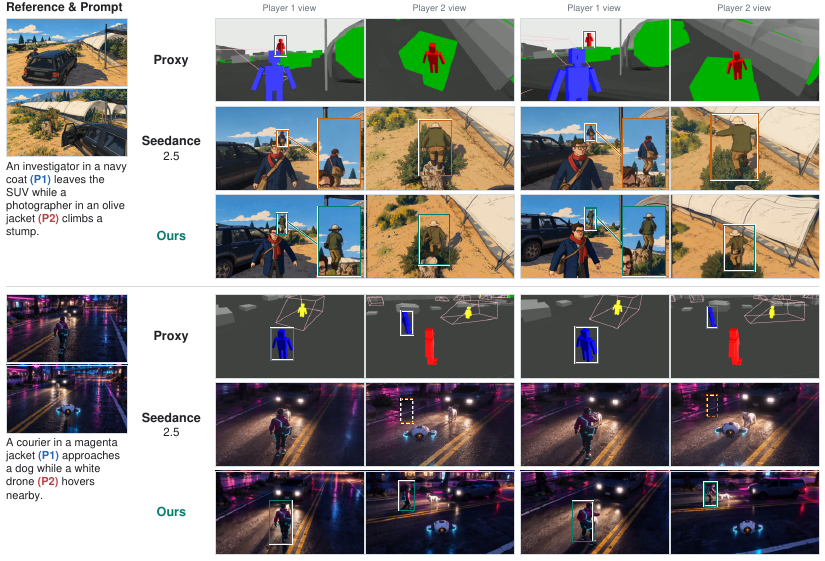}
    \caption{Comparison with Seedance~2.5 across two player views. Boxes highlight the compared player regions.}
    \label{fig:dual-player-consistency}
\end{figure}

\FloatBarrier
\paragraph{Qualitative Cross-View Results.}
Figure~\ref{fig:dual-player-consistency} compares appearance and visibility across views.
Seedance assigns inconsistent clothing to the same player in the farm scene and omits the courier from one view in the night-street scene.
Our renderer maintains consistent clothing in the former and shows the courier in both views in the latter.

\subsection{Proxy Representation Comparison}
\label{sec:proxy_comparison}

\begin{figure}[t]
    \centering
    \includegraphics[width=0.90\linewidth]{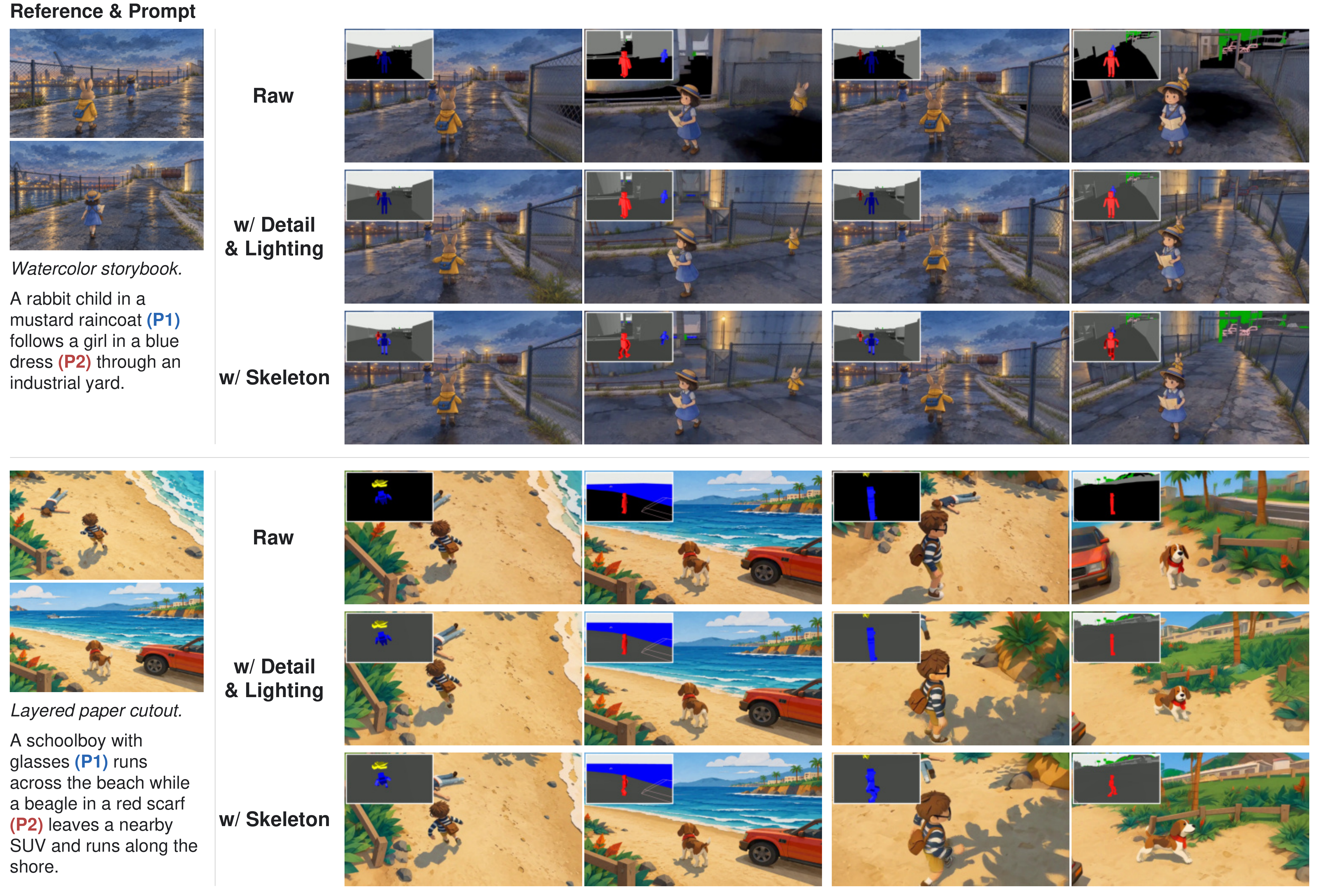}
    \caption{Effect of proxy construction. Rows compare Raw, Detail \& Lighting, and Skeleton, with the corresponding proxy shown in each inset.}
    \label{fig:proxy-construction-ablation}
\end{figure}

Figure~\ref{fig:proxy-construction-ablation} examines three proxy constructions.
Raw provides coarse scene and entity geometry.
Detail \& Lighting adds scene detail and shading, while Skeleton additionally exposes approximate character articulation.
Raw's coarse geometry cannot accurately represent depth relationships, occlusion, and local scene structure in complex environments.
The resulting ambiguity is accompanied by blackened image regions, visible around the industrial structures, and weaker following of the proxy's spatial layout and motion.
Adding scene detail and lighting provides more informative contours, depth cues, and shading, improving the rendering of complex surroundings while retaining coarse character control.

Adding Skeleton makes character pose and limb trajectories more explicit, but this extra precision can overconstrain the video model.
It reduces the freedom of the learned video prior to infer natural articulation from the reference appearance and coarse motion.
In the corresponding videos, characters tend to copy the proxy's prescribed limb movements too literally, producing rigid, less natural motion.
This limitation is particularly apparent when a human-like skeleton controls a non-human reference character, such as the beagle in the beach example.
Based on these observations, we favor detailed environment geometry and lighting with coarse character motion, leaving fine articulation to the video prior.

\FloatBarrier
\subsection{Generation Method Ablation}
\label{sec:generation_ablation}

We compare our joint generation method with a dual-branch variant inspired by MetaWorld~\citep{metaworld}.
The variant assigns a separate generation branch to each player and uses cross-attention to exchange features between the branches during denoising.
Our method generates both views on a concatenated canvas within a shared attention context.

\begin{figure}[!htbp]
    \centering
    \includegraphics[width=0.90\linewidth]{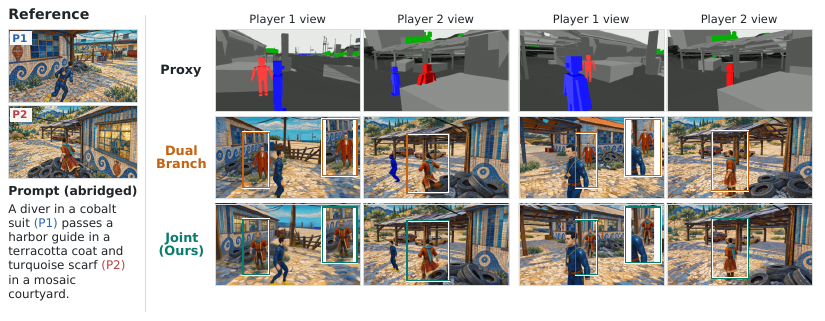}
    \caption{Dual Branch and Joint (Ours) on a mosaic-courtyard case at two matched times. Both methods use step-2,000 checkpoints. Outlines and enlarged crops highlight player~B in player~A's view for comparison with player~B's appearance in their own view.}
    \label{fig:dual-branch-ablation}
\end{figure}

Figure~\ref{fig:dual-branch-ablation} compares the two methods on a shared courtyard scene, focusing on the same player's clothing across views.

\FloatBarrier
\section{Discussion and Conclusion}
\label{sec:conclusion}

\subsection{Conclusion}

Shared 3D proxies provide a spatial interface between a multiplayer environment and a generative video renderer.
Our implementation combines paired camera projections with joint appearance conditioning, adapting a pretrained transformer to synthesize two player views within one context.
The evaluation shows higher DOVER, LAION aesthetic quality, edge alignment, and cross-view player-identity consistency than Seedance~2.5 on the recorded appearance-transfer cases, together with remaining adjacent-frame and reference-preservation errors.

The shared proxy interface could support collaborative virtual environments, with future device--edge--cloud deployment guided by the AI Flow framework~\citep{an2025aiflowperspectivesscenarios,shao2025aiflownetworkedge}.
Extending the renderer to more players requires managing the growing attention context; learned state transitions and sustained interaction remain directions for future work.

\subsection{Limitations}
\label{sec:limitations}

Our current implementation is limited to two-player generation.
The joint generation formulation can in principle extend to $N$ players by concatenating their views within the same context.
At fixed per-view resolution and clip length, however, each additional player increases the token sequence and hence the attention computation and memory requirements of the single-stream diffusion transformer (DiT).
Scaling to larger player groups therefore requires a more efficient way to manage this shared context.

Our current renderer uses bidirectional attention with 50 denoising steps.
Distilling it to eight steps retains good visual quality in our experiments.
Although the proxy can accept inputs in real time, meeting our real-time rendering target requires reducing generation to one or two denoising steps.
Our preliminary attempts show that preserving generation quality under such aggressive distillation remains challenging.

\section{Acknowledgment}
\label{sec:acknowledgment}

We thank Teng Hu for providing advice on the project. We also thank Xinya Wang and Jiahao Yin for collecting the data.

\bibliographystyle{plainnat}
\bibliography{main}

\end{document}